\pdfoutput=1
\documentclass[pdflatex,sn-nature]{sn-jnl}
\usepackage{graphicx}
\usepackage{amsmath,amssymb}
\usepackage{booktabs,tabularx}
\usepackage{placeins}
\usepackage{xcolor}
\usepackage{url}
\unnumbered
\setcitestyle{super,open={},close={}}

\begin{document}
\title[Surface-volume learning for imaging genetics]{Surface-volume self-supervised representation learning of brain MRI for genetic discovery }
\author[1,2]{\fnm{Tian} \sur{Xia}}
\equalcont{These authors contributed equally to this work.}
\author[3]{\fnm{Nuo} \sur{Chen}}
\equalcont{These authors contributed equally to this work.}
\author[3]{\fnm{Zihao} \sur{Zhu}}
\author[4]{\fnm{Huiwen} \sur{Han}}
\author[1,2]{\fnm{Ziqian} \sur{Xie}}
\author*[3]{\fnm{Zhiwen} \sur{Fan}}
\author*[1,2]{\fnm{Degui} \sur{Zhi}}
\email{degui.zhi@yale.edu}
\email{zhiwenfan@tamu.edu}
\affil[1]{\orgdiv{D. Bradley McWilliams School of Biomedical Informatics}, \orgname{The University of Texas Health Science Center at Houston}, \orgaddress{\city{Houston}, \country{US}}}
\affil[2]{\orgdiv{Department of Biomedical Informatics and Data Science}, \orgname{Yale School of Medicine}, \orgaddress{\city{New Haven}, \country{US}}}
\affil[3]{\orgdiv{Department of Electrical and Computer Engineering}, \orgname{Texas A\&M University}, \orgaddress{\city{College Station}, \country{US}}}
\affil[4]{\orgdiv{Department of Psychiatry and Behavioral Sciences}, \orgname{Stanford University School of Medicine}, \orgaddress{\city{Stanford}, \state{CA}, \country{USA}}}

\abstract{
Existing genome-wide association studies (GWAS) of brain imaging provide predefined or deep-learning-derived imaging phenotypes, yet these phenotypes come from either volumetric scans or cortical surface meshes, so each captures only part of the heritable variation in brain anatomy. Here we introduce MEVA (Mesh-Enhanced Volumetric Autoencoder), a self-supervised framework that encodes voxel-level image intensity together with cortical mesh geometry, including curvature and cortical thickness at each surface vertex, into one shared set of imaging features. Combining the mesh and volumetric inputs in MEVA yields modest performance gains in age and sex prediction over models that use either input alone. When these features serve as phenotypes for GWAS in the UK Biobank, they reveal more genome-wide significant loci than features learned from volumes alone or from meshes alone. These results suggest that adding cortical surface geometry to volumetric self-supervised learning captures additional heritable variation and so increases the number of loci detected.}

\maketitle

\section{Introduction}\label{sec:introduction}

Self-supervised learning has become a central tool for extracting structured representations from unlabeled data in domains ranging from natural language to biomedical imaging. In neuroimaging, autoencoders and contrastive learning frameworks trained on brain MRI learn compact representations that capture structural variation across a population without predefined anatomical atlases or expert-curated features~\cite{patel2024unsupervised,dufournet2025self}. These learned representations, sometimes termed \emph{imaging endophenotypes}, are appealing targets for genetic association studies, as they are high-dimensional representations learned directly from the images and may therefore be sensitive to biological variation beyond what hand-crafted imaging-derived phenotypes (IDPs) encode~\cite{patel2024unsupervised,rakowski2024transfergwas}.

Recent advances in machine learning offer flexible methods that learn salient features directly from raw imaging data such as T1-weighted brain MRI, capturing fine-grained structural detail and producing data-driven representations of the underlying anatomy~\cite{giger2018machine}. A growing body of work has applied deep learning to derive image features for genetic discovery, spanning supervised approaches that rely on disease labels or genetic signals~\cite{ning2018classifying,zhou2019effective,li2021method,chakraborty2023deep,zhou2023attentive,li2025imaging,wang2025integration} and unsupervised or self-supervised methods that learn representations without annotation. In the supervised setting, Yu et al.~\cite{yu2024novel} used single-nucleotide polymorphism (SNP) genotypes as class labels to train convolutional neural networks (CNNs) on 2D brain MRI slices, which amounts to a non-linear genome-wide association study (GWAS) but sacrifices interpretability and requires a separate classifier for each of more than 300{,}000 SNPs. In the unsupervised setting, Patel et al.~\cite{patel2024unsupervised} introduced Unsupervised Deep learning derived Imaging Phenotypes (UDIPs), compact 128-dimensional representations learned by 3D convolutional autoencoders directly from raw brain MRI, and GWAS of UDIPs showed higher heritability than most conventional IDPs and uncovered loci absent from earlier IDP-based GWAS. Subsequent work replaced the convolutional encoder with a vision transformer, which captures both local and long-range structure in the image and revealed further genetic associations~\cite{islam2026vision}. Rakowski et al.~\cite{rakowski2024transfergwas} took a transfer-learning approach, applying models pretrained on brain MRI from the Alzheimer's Disease Neuroimaging Initiative (ADNI) or on natural images (ImageNet) to UK Biobank T1-weighted scans to derive low-dimensional representations for GWAS. All of these volumetric approaches, however, operate on 3D intensity arrays alone and leave the folded geometry of the cortical surface unmodeled.

Compared with the growing volumetric literature, the use of mesh- and shape-based deep learning for genetic discovery remains sparse. Bonazzola et al.~\cite{bonazzola2024unsupervised} applied unsupervised ensemble-based phenotyping to 3D surface meshes of the left ventricle and showed that learned mesh phenotypes yield genetic loci beyond those discoverable from handcrafted cardiac traits such as volume and mass. In the brain, Dufournet et al.~\cite{dufournet2025self} trained a self-supervised contrastive model on sulcal representations of cortical folding and identified genetic influences on folding patterns of the anterior cingulate cortex. In a complementary direction, a line of work has treated geometric shape descriptors themselves as multivariate genetic phenotypes. Ge et al.~\cite{ge2016multidimensional} extended heritability estimation to multidimensional traits and showed that shape descriptors of subcortical structures are significantly heritable above and beyond volume. Building on this idea, Primus et al.~\cite{primus2025beyond} used the Laplace--Beltrami spectrum to encode the geometry of 22 subcortical brain structures and the cerebellum and performed multivariate GWAS, identifying 80 unique variants influencing brain shape after controlling for volume and surface area. At the cortical surface, Naqvi et al.~\cite{naqvi2021shared} performed a multivariate GWAS of hierarchically segmented cortical surface morphology in 19{,}644 individuals, identifying 472 loci influencing brain shape, 76 of which were also associated with face shape. Shadrin et al.~\cite{shadrin2021vertex} applied the multivariate omnibus statistical test (MOSTest) to vertex-wise cortical surface area and thickness in 35{,}657 UK Biobank participants and identified 780 unique loci, more than an eightfold increase over univariate region-of-interest GWAS. Beyond the brain itself, Goovaerts et al.~\cite{goovaerts2023joint} conducted a multivariate GWAS of 3D cranial vault shape and identified 30 loci, several of them shared with face and brain shape. Despite these efforts, to our knowledge no prior work has used a self-supervised framework to derive mesh-based brain representations for genetic discovery.

A recurring finding across organs and imaging modalities is that genetic associations from volumetric and surface-derived phenotypes overlap only in part and otherwise diverge substantially. In the heart, Bonazzola et al.~\cite{bonazzola2024unsupervised} found that of 49 study-wide significant loci identified by their deep mesh framework, only 9 had been reported in prior GWAS of left-ventricular phenotypes, and 20 were attributable exclusively to the mesh-based representation. Burns et al.~\cite{burns2024genetic} similarly reported 43 loci from principal components of bi-ventricular surface meshes, 14 of which had not previously been associated with cardiac traits. In the brain, Primus et al.~\cite{primus2025beyond} showed that, after controlling for volume and surface area, 49 of 80 shape-associated variants had previously been linked to brain shape-related traits, whereas the remaining 31 were novel, confirming shape-specific signal beyond gross volumetric measures. Van der Meer et al.~\cite{vandermeer2021genetic} found substantial genetic overlap between sulcal depth and other cortical traits such as thickness and surface area, yet sulcal depth still showed a higher locus yield and higher heritability. Naqvi et al.~\cite{naqvi2021shared} reported that a multivariate GWAS of cortical surface shape recaptured part of the signal from prior surface-area and subcortical-volume GWAS (121 of 472 loci), whereas most loci (351) were newly described. The consistent pattern across these studies is that volumetric measures capture coarse size variation, whereas mesh and shape representations encode geometry that is at least partly genetically independent. Most comparisons to date, however, rest on catalog-based overlap counts after covariate adjustment and stop short of formal colocalization or fine-mapping, so the mechanistic relationship between the two signal sources remains incompletely resolved. This partial complementarity motivates a framework that captures both sources of heritable variation at once.

We address this gap with the Mesh-Enhanced Volumetric Autoencoder (MEVA), a dual-stream architecture that uses volumetric MRI and cortical surface representations together. The volumetric stream encodes the T1-weighted MRI volume, whereas the mesh stream separately encodes vertex-wise cortical features derived from FreeSurfer-reconstructed surfaces~\cite{fischl2012freesurfer}. MEVA encodes each modality on its own and then combines the two representations through a residual fusion step.

We first trained and evaluated the two streams separately. GWAS of each stream showed that the mesh representation identified about two thirds as many significant loci as the volumetric representation (108 versus 156), yet the loci found by the two streams overlapped only in part. These findings suggest that the mesh representation carries weaker genetic signal overall but captures complementary genetic information that the volumetric stream only partly represents.

This observation motivated residual fusion over direct feature fusion. With the volumetric representation carrying stronger genetic signal, unconstrained fusion risks either letting the volumetric stream dominate the joint representation and suppress mesh-specific information, or perturbing the strong volumetric representation and degrading genetic discovery. We therefore treat the volumetric representation as the primary one and learn the contribution of the mesh stream as a residual correction on top of it. This design preserves the genetic information already captured by the volumetric stream while allowing complementary mesh-derived information to improve the representation where it is informative. In the limiting case where the mesh stream adds little, the fused representation falls back toward the volumetric representation and performance is largely retained.

We evaluate MEVA in three stages. First, we assess how much phenotypic information the learned representations carry through prediction of age and sex as well as mesh-derived IDPs, comparing MEVA with volumetric-only and mesh-only baselines to quantify the benefit of integration. Second, we perform GWAS on each dimension of the learned representation in the UK Biobank~\cite{miller2016multimodal} discovery cohort, using the same individuals and pipeline for all models so that differences in locus counts and identity reflect the representation alone. Third, we will carry out downstream genetic analyses (enrichment of gene sets and tissue expression, together with genetic correlation with clinical phenotypes) to establish the biological relevance of MEVA-specific discoveries, and we will examine the encoder to learn where and why the mesh stream contributes new signal.

This paper makes three contributions:
\begin{itemize}
    \item We demonstrate that cortical mesh features learned with a self-supervised autoencoder carry predictive and genetic signal complementary to that of volumetric representations.
    \item We show that MEVA, by combining both streams, increases the number of genome-wide significant loci beyond either single-modality baseline.
    \item We will provide an explainability analysis to test whether the mesh and volumetric streams contribute to distinct genetic axes, which would explain why the combined representation discovers more loci.
\end{itemize}

\section{Results}\label{sec:results}

\subsection{Phenotypic information in cortical mesh representations}
The first analysis assesses whether cortical geometry provides an informative representation before integration with volumetric MRI. A mesh-only autoencoder encodes the mid-thickness vertex coordinates, cortical thickness and curvature of FreeSurfer-derived cortical surfaces. Mesh-only phenotypes predicted age with a Pearson r of 0.676 and sex with an AUC of 0.927 (Table~\ref{tab:comparison}).

\begin{figure}[!htbp]
\centering
\includegraphics[width=\linewidth]{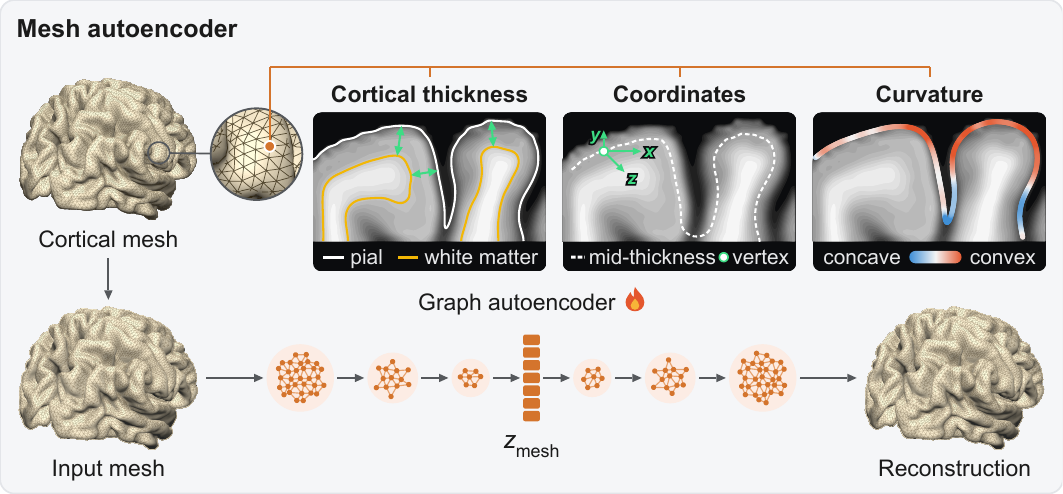}
\caption{\textbf{Mesh-only autoencoder.} A graph autoencoder encodes the fsaverage4 cortical mesh of each hemisphere into the embedding $z_{\mathrm{mesh}}$, from which the vertex features are reconstructed. Each vertex carries its mid-thickness $x$, $y$ and $z$ coordinates, cortical thickness (the distance between the pial and white-matter surfaces) and curvature, from concave (sulcal fundi) to convex (gyral crowns).}\label{fig:mesh}
\end{figure}

The mesh autoencoder is outlined in Fig.~\ref{fig:mesh}.

\begin{table}[!htbp]
\caption{\textbf{Comparison of the learned representations.}}\label{tab:comparison}
\small
\setlength{\tabcolsep}{3pt}
\begin{tabular*}{\textwidth}{@{\extracolsep{\fill}}lccccccc@{}}
\toprule
 & \multicolumn{2}{c}{Prediction} & \multicolumn{3}{c}{Cortical $\Delta R^2$ $\uparrow$} & \multicolumn{2}{c}{Genetics}\\
\cmidrule(lr){2-3}\cmidrule(lr){4-6}\cmidrule(l){7-8}
Representation & Age Pearson r $\uparrow$ & Sex AUC $\uparrow$ & Thickness & Area & Volume & Loci $\uparrow$ & $\sum h^2_{\mathrm{SNP}}$ $\uparrow$\\
\midrule
Volume-only & 0.768 & 0.976 & 0.091 & 0.258 & 0.209 & 156 & 33.7\\
Mesh-only & 0.676 & 0.927 & \textbf{0.580} & \textbf{0.344} & \textbf{0.376} & 108 & $26.4$\\
MEVA & \textbf{0.774} & \textbf{0.986} & 0.393 & 0.312 & 0.310 & \textbf{168} & \textbf{36.3}\\
\botrule
\end{tabular*}
\footnotetext{All representations have 128 phenotypes and are evaluated in the same 35,298 held-out UK Biobank participants; bold marks the best value in each column. Age mean Pearson r and sex AUC are from five-fold cross-validation. Cortical information is the variance of FreeSurfer regional thickness, surface area and grey-matter volume explained beyond covariates, averaged over the 68 Desikan--Killiany regions. Bootstrap 95\% confidence intervals lie within $\pm0.006$ of every sex AUC and $\Delta R^2$ value; Loci are from TorchGWAS analysis of 128 phenotypes, followed by the minimal p value of aggregating multivariate summary statistics and FUMA for loci aggregation; $\sum h^2_{\mathrm{SNP}}$ is the summed SNP heritability from the sum-h2 pipeline (Methods). }
\end{table}
\subsection{Integrating cortical geometry and volumetric intensity}
\begin{figure}[!htbp]
    \centering
    \includegraphics[width=\linewidth]{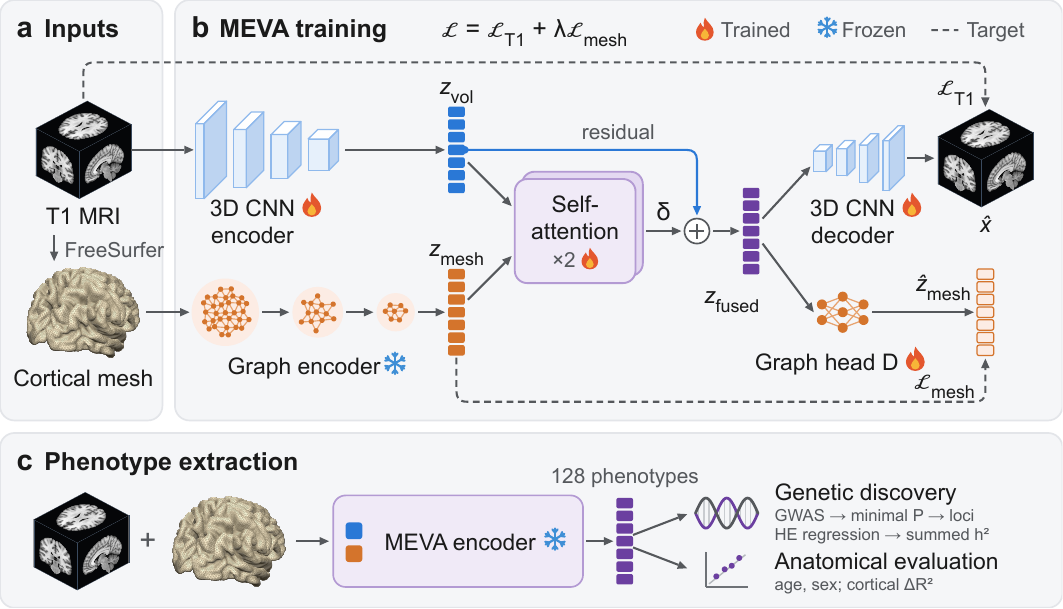}
    \caption{\textbf{MEVA architecture, training objective and phenotype extraction.}
\textbf{a}, A T1-weighted MRI volume and its FreeSurfer-reconstructed cortical surface are used as inputs.
\textbf{b}, A 3D CNN encodes the volume into $z_{\mathrm{vol}}$, while a pretrained, frozen mesh autoencoder encodes the surface into $z_{\mathrm{mesh}}$. A self-attention block jointly processes the embeddings, and its output is projected to a residual $\delta$ that is added to $z_{\mathrm{vol}}$ to obtain $z_{\mathrm{fused}}\in\mathbb{R}^{128}$. Training uses T1 reconstruction and mesh-embedding prediction losses, $\mathcal{L}=\mathcal{L}{\mathrm{T1}}+\lambda\mathcal{L}{\mathrm{mesh}}$ ($\lambda=10$).
\textbf{c}, The trained encoder generates 128 imaging phenotypes from each participant. These phenotypes are used for genetic discovery and anatomical evaluation, including age, sex and cortical $\Delta R^2$ prediction.}
    \label{fig:integration}
\end{figure}

The second analysis compares MEVA (architecture in Fig.~\ref{fig:integration}) with volume-only and mesh-only autoencoders on the same prediction tasks and cortical measures (Table~\ref{tab:comparison}). MEVA phenotypes achieved a Pearson correlation of $r=0.774$ with age, compared with $r=0.768$ for volume-only and $r=0.676$ for mesh-only phenotypes, and predicted sex with an AUC of 0.986, compared with 0.976 and 0.927, respectively. 

\subsection{Genetic discovery from multimodal representations}
\begin{figure}[!htbp]
    \centering
    \begin{minipage}[t]{0.29\linewidth}\raggedright\textbf{a}\\[2pt]\includegraphics[width=\linewidth]{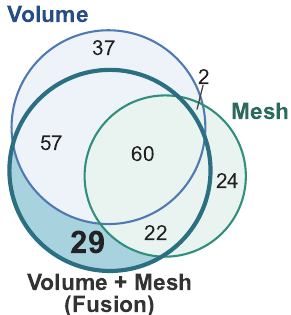}\end{minipage}\hfill
    \begin{minipage}[t]{0.68\linewidth}\raggedright\textbf{b}\\[2pt]\includegraphics[width=\linewidth]{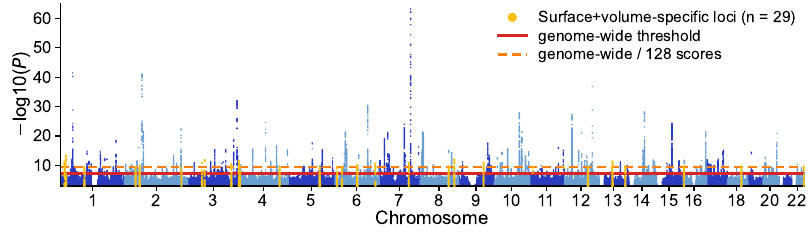}\end{minipage}
    \caption{\textbf{Genetic associations across imaging representations.} \textbf{a}, Overlap of genome-wide significant loci identified by the volume-only (156 loci), mesh-only (108 loci), and MEVA (168 loci; shown as Volume + Mesh (Fusion)) representations; loci were merged across models when their boundaries, padded by 125\,kb on each side, overlapped. Twenty-nine loci were identified only by MEVA.
    \textbf{b}, Manhattan plot of the MEVA minimum-P GWAS ($N = 35{,}298$). SNPs within the 29 MEVA-specific loci (surface+volume-specific loci in the legend) are highlighted in yellow.}
    \label{fig:gwas}
\end{figure}

The third analysis compares genetic associations across representations in a common UK Biobank discovery cohort, using the same participants, covariates and association pipeline for every representation (Methods, Genome-wide association analysis). Combining the per-phenotype summary statistics with the minimum-P method, followed by FUMA locus clumping, MEVA yielded 168 loci, compared with 156 for the volume-only and 108 for the mesh-only representation, and it had the highest summed SNP heritability (36.3, versus 33.7 and 26.4; Table~\ref{tab:comparison}).
Because locus count alone do not show what fusion adds, we compared the loci across representations, treating two loci as the same when their intervals, padded by 125 kb on each side, overlapped on the same chromosome (Fig.~\ref{fig:gwas}a; Methods, Locus overlap analysis). Of the 168 MEVA loci, 29 were not detected with either the volume-only or the mesh-only representation.

The genetic association analysis is organized in Fig.~\ref{fig:gwas}, with a summary in Table~\ref{tab:comparison}.

\section{Discussion}
\label{sec:discussion}

In this work, we introduced MEVA, a dual-stream autoencoder that augments a volumetric MRI representation with a residual contribution from cortical mesh features, and showed that this integration yields a representation that is both more phenotypically informative and more genetically discoverable than either modality alone. MEVA achieved the best performance on both phenotype prediction tasks, reaching a Pearson correlation of 0.774 for brain age (compared with 0.768 for the volume-only and 0.676 for the mesh-only representation) and an AUC of 0.986 for sex classification (compared with 0.976 and 0.927, respectively). Beyond improved prediction, MEVA identified 168 genome-wide significant loci, exceeding both the volume-only (156 loci) and mesh-only (108 loci) baselines, and 29 of these loci were detected by neither unimodal representation. MEVA also attained the highest summed SNP heritability ($\sum h^2_{\mathrm{SNP}} = 36.3$, compared with 33.7 and 26.4)~\cite{xia2026sumh2}, indicating that the additional loci reflect a genuine gain in captured heritable variation rather than a redistribution of existing signal across latent dimensions. Notably, although the mesh-only representation carried the weakest genetic signal on its own, adding it as a residual to the volumetric stream substantially increased the cortical thickness variance explained ($\Delta R^2 = 0.393$ versus $0.091$ for the volume-only model), showing that the fused phenotypes retain cortical information supplied by the mesh stream. Together, these results support residual fusion as an effective strategy for combining a strong primary modality with a weaker but complementary one, preserving the genetic signal of the volumetric representation while extending it with mesh-derived information.

Residual fusion is only one of several strategies for integrating volumetric and surface representations, and alternative designs may capture complementary signal in different ways. A first alternative is contrastive alignment, in which modality-specific encoders are trained to map paired views of the same individual to nearby points in a shared embedding space~\cite{radford2021learning}. We previously showed that contrastive learning across brain imaging modalities yields heritable multimodal representations~\cite{xia2026learning}, and extending this objective to volume--mesh pairs could encourage the model to emphasize anatomical variation that is consistently expressed in both modalities. However, because contrastive objectives reward shared information, they may attenuate precisely the mesh-specific signal that residual fusion is designed to preserve. A second alternative is allocation-based fusion, in which the latent budget is explicitly partitioned into shared and modality-specific dimensions, or the contribution of each modality is adaptively weighted per dimension. Such a design would make the division of labor between the volumetric and mesh streams explicit, and would allow loci to be attributed to shared or modality-specific axes directly rather than through post hoc interpretation. A third alternative is fine-grained cross-attention between local tokens of the two modalities~\cite{vaswani2017attention,chen2021crossvit}, for example between 3D patches of the MRI volume and vertices or vertex patches of the cortical mesh. Because each cortical vertex has a well-defined spatial location in the MRI volume, cross-attention at the patch and vertex level could learn anatomically grounded correspondences and allow surface geometry to modulate the volumetric features of the underlying tissue, at the cost of substantially higher computational and memory demands. Systematically comparing these fusion strategies under the same GWAS pipeline and evaluation framework~\cite{xia2026sumh2} will be an important direction for determining which inductive biases best expose heritable brain variation.

The present analyses indicate that MEVA improves both phenotypic prediction and locus discovery, but they do not yet characterize the biology underlying the MEVA-specific loci. We are therefore extending this work in two directions. First, we are performing downstream genetic analyses of the MEVA GWAS, including gene-based and gene-set enrichment with MAGMA~\cite{deleeuw2015magma}, tissue-expression enrichment using GTEx~\cite{gtex2020atlas}, and genetic correlation with neurological, psychiatric, and cardiometabolic phenotypes~\cite{bulik2015atlas}, to determine whether the loci gained through fusion map to neurobiologically coherent pathways. Second, we are developing an explainability analysis that tests whether the fusion model's attributions recover genetically informed cortical topography at the 29 fusion-specific lead variants. This analysis compares decoder-perturbation and input-gradient maps derived from the MEVA encoder with gene-expression-weighted cortical profiles constructed from the Allen Human Brain Atlas~\cite{hawrylycz2012anatomically}, weighted by MAGMA gene associations, GTEx brain eQTL effects, or brain transcriptome-wide association statistics~\cite{barbeira2018exploring}. Concordance between the two would indicate that the mesh stream contributes signal that is anatomically localized to the cortical regions where the implicated genes are expressed, providing a mechanistic account of why the integrated representation discovers loci that neither unimodal baseline detects.

\section{Methods}\label{sec:methods}

\subsection{UK Biobank participants}
The UK Biobank \cite{sudlow2015uk} is a nationwide population resource encompassing approximately 500,000 participants recruited across the United Kingdom, with baseline phenotypic and genetic data collected between 2006 and 2010 and brain imaging from 2014 onwards. The study operates under established ethical approval, with governance and oversight details available at \url{https://www.ukbiobank.ac.uk/learn-more-about-uk-biobank/governance/ethics-advisory-committee}. Models were developed in the quality-controlled subset of participants described below. Phenotypes were then extracted for 35,298 participants with T1-weighted MRI, cortical surfaces and imputed genotypes~\cite{bycroft2018uk} who were not used for model training; these participants form the genetic analysis cohort for every representation. The genetic analysis cohort had a mean age at imaging of 67.8 years (s.d. 7.7; range 46--87).

\subsection{MRI acquisition and volumetric preprocessing}
UK Biobank MRIs were downloaded on October 15, 2021. UK Biobank participants had birth years between 1934 and 1971. UK Biobank has provided a bias-field-corrected version of the brain-extracted T1-weighted (T1) acquired on Siemens Skyra 3T scanners (software VD13A SP4 as of October 2015) with a standard Siemens 32-channel RF receive head coil, at 1-mm isotropic resolution (\url{https://biobank.ctsu.ox.ac.uk/crystal/crystal/docs/brain_mri.pdf}).

To promote generalizability and minimize manual feature engineering, we adopted the standard preprocessing pipeline developed by the UK Biobank MRI team. Brain MRI preprocessing was performed primarily using FSL (\url{https://www.fmrib.ox.ac.uk/ukbiobank/}), and included defacing, brain extraction with BET, linear and non-linear registration to standard MNI152 space using FLIRT and FNIRT, and bias-field correction using FAST. Bias-field-corrected, brain-extracted T1-weighted images were retained for analysis. All images were subsequently linearly registered to MNI152 space using affine transformation with 12 degrees of freedom via UK Biobank-provided precomputed matrices in FLIRT. Linear registration was chosen to normalize head size and achieve cross-subject alignment while preserving subject-specific structural deformations, in contrast to non-linear registration. The resulting linearly registered, defaced, bias-field-corrected images were used in all downstream analyses. Image intensities were normalized on a per-scan basis using Z-score normalization. Following affine registration, T1 weighted volumes had dimensions of 182 $\times$ 218 $\times$ 182 voxels and were zero-padded to 182 $\times$ 224 $\times$ 182. %

To construct a high-quality deep learning dataset, we leveraged UK Biobank's precomputed MRI quality metrics: the inverted contrast-to-noise ratio (Data Field 25735) and the discrepancy between T2 FLAIR and T1 brain images (Data Field 25736). Images with values below the 95th percentile for both metrics (corresponding to higher-quality scans) were retained. For participants with multiple visits, only the first scan was kept to ensure consistency across the dataset. This filtering resulted in a cohort of 6,130 scans from individuals of diverse ethnic backgrounds. The pretrained volumetric autoencoder was trained on a random split of these scans into 4,597 training (75\%) and 1,533 validation (25\%) images, with the validation set reserved for hyperparameter tuning and checkpointing. The mesh autoencoder and MEVA were trained on the 6,067 of these participants with usable cortical surfaces.

\subsection{Cortical surface preparation}
Cortical surfaces were reconstructed with FreeSurfer%
~and resampled to the fsaverage4 icosahedral mesh, giving 2,562 vertices per hemisphere. Each vertex carries five channels: the $x$, $y$ and $z$ coordinates of the mid-thickness surface, cortical thickness and curvature. The coordinates of each hemisphere were centred and scaled with a single centre and scale estimated on the training set, and thickness and curvature were standardized with training-set means and standard deviations.

\subsection{Mesh autoencoder}
The mesh-only encoder is a graph autoencoder with 2.9 million parameters, applied with shared weights to the fsaverage4 mesh graph of each hemisphere (2,562 vertices). The three coordinate channels and the two morphometric channels are embedded separately by two-layer perceptrons into 32 channels and summed. Four graph convolution layers with 32, 64, 64 and 32 output channels are interleaved with fixed icosahedral decimation, which reduces each hemisphere from 2,562 to 642, 162 and 42 vertices. The 42-vertex features of the two hemispheres are concatenated and mapped through a two-layer perceptron (512 hidden units) to the 128-dimensional embedding $z_{\mathrm{mesh}}$. The decoder maps $z_{\mathrm{mesh}}$ back to 42 vertices per hemisphere and mirrors the encoder with four graph convolution layers, upsampling between levels by averaging the two parent vertices of each new vertex, and a vertex-wise linear layer reconstructs the five input channels. The reconstruction target is each vertex's deviation from the training mean, with each channel scaled to unit training variance, and the loss is the mean squared error over cortical vertices. The model was trained on 4,853 participants, with 607 for validation and 607 held out, with a learning rate of $10^{-4}$ and a batch size of 32 for 500 epochs; the checkpoint with the lowest validation reconstruction error (epoch 498) was retained. No genetic information was used.

\subsection{MEVA fusion and training}
MEVA combines the pretrained volumetric autoencoder with the frozen mesh encoder (Fig.~\ref{fig:integration}). A fusion module $f$ computes a residual $\delta=f(z_{\mathrm{vol}},z_{\mathrm{mesh}})$, and the fused phenotypes are
\begin{equation}
z_{\mathrm{fused}}=z_{\mathrm{vol}}+\delta .
\end{equation}
The last layer of $f$ is initialized to zero, so training begins from the volumetric representation, and $z_{\mathrm{fused}}$ is passed to the convolutional decoder to reconstruct the T1 image. Training used 5,461 participants, with 606 for validation and a batch size of two, in two stages. In stage 1 (4 epochs), $f$ is a residual multilayer perceptron applied to $[z_{\mathrm{vol}},z_{\mathrm{mesh}}]$ (layer normalization, linear $256\to128$, GELU, linear $128\to128$), and the loss combines masked T1 reconstruction with an instance-discrimination InfoNCE loss (weight 0.1, temperature 0.1, queue of 512) whose positive pairs are fused embeddings of two augmented views of the same T1 image. The convolutional encoder was fine-tuned with a learning rate of $5\times10^{-5}$ and the fusion module with $10^{-3}$. In stage 2 (6 epochs), the encoder and a learned mesh adapter are retained and $f$ is replaced by a two-token transformer: $z_{\mathrm{vol}}$ and the adapted $z_{\mathrm{mesh}}$ are processed as two tokens by two layers of four-head self-attention, concatenated and projected to $\delta$. The InfoNCE term is removed, and a distillation head $D$ (linear $128\to128$, GELU, linear $128\to128$) predicts $z_{\mathrm{mesh}}$ from $z_{\mathrm{fused}}$ with the target under stop-gradient,
\begin{equation}
\mathcal{L}=\mathcal{L}_{\mathrm{T1}}+\lambda\,\mathcal{L}_{\mathrm{mesh}},\qquad \lambda=10 .
\end{equation}
The final epoch of each stage was used. Surface information therefore enters MEVA only through $z_{\mathrm{mesh}}$ and the distillation loss; vertex features are not reconstructed. The modules added to the volumetric autoencoder (adapter, fusion block and distillation head) have about 0.18 million parameters (Supplementary Table~\ref{tab:capacity}).

\subsection{Volumetric autoencoder baselines and training}
The volumetric stream uses the CNN autoencoder of our previous work~\cite{patel2024unsupervised}. The volume-only representation in this study is the pretrained CNN autoencoder (138.1 million parameters, trained for 40 epochs), used without further training; it also initializes the MEVA volume encoder.

Briefly, the CNN architecture is an autoencoder with CNN blocks of $3\times3\times3$ kernels and $2\times2\times2$ max pooling. The model consists of 5 encoding CNN blocks, a fully connected layer with dimension 128 and 5 decoding CNN blocks.

\subsection{Embedding extraction}
For each representation, the encoder was frozen and the decoder and any auxiliary heads were removed. The MEVA encoder takes both the T1 image and $z_{\mathrm{mesh}}$ computed from the participant's cortical surface. The volume-only, mesh-only and MEVA encoders were applied to the same 35,298 genetic-analysis participants, giving a $35{,}298\times128$ phenotype matrix per representation.

\subsection{Phenotypic prediction and statistical analysis}
Phenotypes were evaluated in the 35,298 held-out participants. Age was predicted by ridge regression ($\alpha=1$) with five-fold cross-validation and summarized by the Pearson correlation and mean absolute error between predicted and observed age. Sex was predicted by class-balanced logistic regression with stratified five-fold cross-validation and summarized by the AUC. Cortical information was measured as the incremental variance explained, $\Delta R^2=R^2(\text{measure}\sim\text{phenotypes}+\text{covariates})-R^2(\text{measure}\sim\text{covariates})$, from ordinary least squares fitted in sample and averaged over the components of each measure set: FreeSurfer regional thickness, surface area and grey-matter volume of the 68 Desikan--Killiany regions~\cite{desikan2006automated}. Confidence intervals were obtained from 500 bootstrap resamples of participants; for $\Delta R^2$, both models were refitted on every resample and the interval is the point estimate $\pm1.96$ bootstrap standard errors.

\subsection{Genotype quality control}
We applied the quality-control pipeline of our previous work \cite{patel2024unsupervised} to the imputed autosomal genotypes. Duplicated variants were removed, and individuals whose genetically inferred sex did not match their self-reported sex were excluded. %

\subsection{Genome-wide association analysis}
For each representation, the 128 phenotypes were tested on their original scale, without rank-based inverse normal transformation, for association with 8,931,083 variants in the 35,298 participants, using TorchGWAS with covariates, including age, age$^2$, sex, their interactions, genetic principal components. %
The per-phenotype summary statistics were combined together with minimal p value, and genome-wide significance was set at $P<5\times10^{-8}/128$. Total SNP heritability across multiple phenotypes was estimated with a sum-h2 pipeline \cite{xia2026sumh2}. Basically, we define the total heritability as $\text{tr}(P^{-1}G)$, and it could be estimated by first using PCA to decorrelate phenotypes, followed by sum up heritability from single phenotype estimated by the Haseman--Elston regression. %

\subsection{Genomic locus annotation}
Genomic loci were annotated using FUMA \cite{watanabe2017functional}. Lead SNPs were first identified based on linkage disequilibrium ($r^2\leq0.1$) and physical proximity ($<$250 kb), and assigned to non-overlapping genomic loci. Within each locus, the SNP with the lowest p-value (the top lead SNP) was used to represent the locus.

\subsection{Locus overlap analysis}
To assess locus overlap, each identified locus was extended by 125 kb on either side, and interval trees were constructed for each chromosome. Loci from our analysis were then queried against these interval trees to determine overlapping regions.

\subsection{Ethics approval and consent}
Our analysis was approved by the UTHealth Houston committee for the protection of human subjects under No. HSC-SBMI-20-1323. UKBB has secured informed consent from the participants in the use of their data for approved research projects. UK Biobank data were accessed via approved project 24247.

\section*{Data availability}
The self-supervised imaging phenotypes and GWAS summary statistics are available upon reasonable request after removing sensitive UK Biobank participant information. Genomic loci annotation used data from FUMA (\url{https://fuma.ctglab.nl/}). Individual data from UKBB can be requested with proper registration at \url{https://www.ukbiobank.ac.uk/}. All unrestricted data supporting the findings are also available from the corresponding author upon request.

\section*{Code availability}
The previously published CNN and ViT model code and checkpoints are publicly available on GitHub at \url{https://github.com/ZhiGroup/DeepENDO}, \url{https://github.com/ZhiGroup/UDIP-ViT}. The github repo of surface mesh model and fusion model is under construction. 

\section*{Acknowledgements}
This work was supported by grants from the National Institute on Aging U01AG070112 and R01AG081398.
This research has been conducted using the UK Biobank Resource under Application Number 24247.
We used a large language model to assist with language editing and clarity of presentation. The authors reviewed and edited all content and take full responsibility for the accuracy and integrity of the work.

\section*{Competing interests}
The authors declare no competing interests.

\section*{Additional information}
Correspondence and requests for materials should be addressed to Degui Zhi (\href{mailto:degui.zhi@uth.tmc.edu}{degui.zhi@uth.tmc.edu}) or Zhiwen Fan (\href{mailto:zhiwenfan@tamu.edu}{zhiwenfan@tamu.edu}).

\begingroup
\setlength{\bibsep}{3pt plus 1pt}
\bibliography{references}

\begin{thebibliography}{10}
\expandafter\ifx\csname url\endcsname\relax
  \def\url#1{\burl{#1}}\fi
\expandafter\ifx\csname urlprefix\endcsname\relax\def\urlprefix{URL }\fi
\providecommand{\bibinfo}[2]{#2}
\providecommand{\eprint}[2][]{\url{#2}}
\providecommand{\doi}[1]{\url{https://doi.org/#1}}
\bibcommenthead

\bibitem{patel2024unsupervised}
\bibinfo{author}{Patel, K.} \emph{et~al.}
\newblock \bibinfo{title}{Unsupervised deep representation learning enables
  phenotype discovery for genetic association studies of brain imaging}.
\newblock \emph{\bibinfo{journal}{Communications Biology}}
  \textbf{\bibinfo{volume}{7}}, \bibinfo{pages}{414} (\bibinfo{year}{2024}).

\bibitem{dufournet2025self}
\bibinfo{author}{Dufournet, A.} \emph{et~al.}
\newblock \bibinfo{title}{A self-supervised learning framework for discovering
  cortical folding patterns under genetic influence: Application to the
  anterior cingulate cortex}.
\newblock \emph{\bibinfo{journal}{Imaging Neuroscience}}
  \textbf{\bibinfo{volume}{3}}, \bibinfo{pages}{IMAG.a.987}
  (\bibinfo{year}{2025}).

\bibitem{rakowski2024transfergwas}
\bibinfo{author}{Rakowski, A.}, \bibinfo{author}{Monti, R.} \&
  \bibinfo{author}{Lippert, C.}
\newblock \bibinfo{title}{{TransferGWAS} of {T1}-weighted brain {MRI} data from
  {UK Biobank}}.
\newblock \emph{\bibinfo{journal}{PLoS Genetics}}
  \textbf{\bibinfo{volume}{20}}, \bibinfo{pages}{e1011332}
  (\bibinfo{year}{2024}).

\bibitem{giger2018machine}
\bibinfo{author}{Giger, M.~L.}
\newblock \bibinfo{title}{Machine Learning in Medical Imaging}.
\newblock \emph{\bibinfo{journal}{Journal of the American College of
  Radiology}} \textbf{\bibinfo{volume}{15}}, \bibinfo{pages}{512--520}
  (\bibinfo{year}{2018}).

\bibitem{ning2018classifying}
\bibinfo{author}{Ning, K.} \emph{et~al.}
\newblock \bibinfo{title}{Classifying {Alzheimer's} disease with brain imaging
  and genetic data using a neural network framework}.
\newblock \emph{\bibinfo{journal}{Neurobiology of Aging}}
  \textbf{\bibinfo{volume}{68}}, \bibinfo{pages}{151--158}
  (\bibinfo{year}{2018}).

\bibitem{zhou2019effective}
\bibinfo{author}{Zhou, T.}, \bibinfo{author}{Thung, K.-H.},
  \bibinfo{author}{Zhu, X.} \& \bibinfo{author}{Shen, D.}
\newblock \bibinfo{title}{Effective feature learning and fusion of
  multimodality data using stage-wise deep neural network for dementia
  diagnosis}.
\newblock \emph{\bibinfo{journal}{Human Brain Mapping}}
  \textbf{\bibinfo{volume}{40}}, \bibinfo{pages}{1001--1016}
  (\bibinfo{year}{2019}).

\bibitem{li2021method}
\bibinfo{author}{Li, Y.}, \bibinfo{author}{Liu, Y.}, \bibinfo{author}{Wang, T.}
  \& \bibinfo{author}{Lei, B.} \bibinfo{title}{A Method for Predicting
  {Alzheimer's} Disease Based on the Fusion of Single Nucleotide Polymorphisms
  and Magnetic Resonance Feature Extraction}.
\newblock In \emph{\bibinfo{booktitle}{Multimodal Learning for Clinical
  Decision Support (ML-CDS 2021), Held in Conjunction with MICCAI 2021}}, Vol.
  \bibinfo{volume}{13050} of \emph{\bibinfo{series}{Lecture Notes in Computer
  Science}}, \bibinfo{pages}{105--115} (\bibinfo{publisher}{Springer},
  \bibinfo{address}{Cham}, \bibinfo{year}{2021}).

\bibitem{chakraborty2023deep}
\bibinfo{author}{Chakraborty, D.} \emph{et~al.}
\newblock \bibinfo{title}{Deep Learning-Based Feature Extraction with {MRI}
  Data in Neuroimaging Genetics for {Alzheimer's} Disease}.
\newblock \emph{\bibinfo{journal}{Genes}} \textbf{\bibinfo{volume}{14}},
  \bibinfo{pages}{626} (\bibinfo{year}{2023}).

\bibitem{zhou2023attentive}
\bibinfo{author}{Zhou, R.} \emph{et~al.} \bibinfo{title}{Attentive Deep
  Canonical Correlation Analysis for Diagnosing {Alzheimer's} Disease Using
  Multimodal Imaging Genetics}.
\newblock In \emph{\bibinfo{booktitle}{Medical Image Computing and Computer
  Assisted Intervention -- MICCAI 2023}}, Vol. \bibinfo{volume}{14221} of
  \emph{\bibinfo{series}{Lecture Notes in Computer Science}},
  \bibinfo{pages}{681--691} (\bibinfo{publisher}{Springer},
  \bibinfo{address}{Cham}, \bibinfo{year}{2023}).

\bibitem{li2025imaging}
\bibinfo{author}{Li, Y.}, \bibinfo{author}{Niu, D.}, \bibinfo{author}{Qi, K.},
  \bibinfo{author}{Liang, D.} \& \bibinfo{author}{Long, X.}
\newblock \bibinfo{title}{An imaging and genetic-based deep learning network
  for {Alzheimer's} disease diagnosis}.
\newblock \emph{\bibinfo{journal}{Frontiers in Aging Neuroscience}}
  \textbf{\bibinfo{volume}{17}}, \bibinfo{pages}{1532470}
  (\bibinfo{year}{2025}).

\bibitem{wang2025integration}
\bibinfo{author}{Wang, Y.} \emph{et~al.}
\newblock \bibinfo{title}{Integration of Genetic and Imaging Data for
  {Alzheimer's} Disease Diagnosis and Interpretation}.
\newblock \emph{\bibinfo{journal}{Advanced Science}}
  \textbf{\bibinfo{volume}{12}}, \bibinfo{pages}{e07629}
  (\bibinfo{year}{2025}).

\bibitem{yu2024novel}
\bibinfo{author}{Yu, S.}, \bibinfo{author}{Wu, J.}, \bibinfo{author}{Shao, Y.},
  \bibinfo{author}{Qiu, D.} \& \bibinfo{author}{Qin, Z.~S.}
\newblock \bibinfo{title}{A novel classification framework for genome-wide
  association study of whole brain {MRI} images using deep learning}.
\newblock \emph{\bibinfo{journal}{PLoS Computational Biology}}
  \textbf{\bibinfo{volume}{20}}, \bibinfo{pages}{e1012527}
  (\bibinfo{year}{2024}).

\bibitem{islam2026vision}
\bibinfo{author}{Islam, S.~R.}, \bibinfo{author}{Xia, T.}, \bibinfo{author}{He,
  W.}, \bibinfo{author}{Xie, Z.} \& \bibinfo{author}{Zhi, D.}
\newblock \bibinfo{title}{Vision transformer autoencoders captures local and
  non-local features in brain imaging to reveal novel genetic associations}.
\newblock \emph{\bibinfo{journal}{Communications Biology}}
  (\bibinfo{year}{2026}).
\newblock \urlprefix\url{https://www.nature.com/articles/s42003-026-10430-6}.
\newblock \bibinfo{note}{Accepted manuscript, published online 9 June 2026}.

\bibitem{bonazzola2024unsupervised}
\bibinfo{author}{Bonazzola, R.} \emph{et~al.}
\newblock \bibinfo{title}{Unsupervised ensemble-based phenotyping enhances
  discoverability of genes related to left-ventricular morphology}.
\newblock \emph{\bibinfo{journal}{Nature Machine Intelligence}}
  \textbf{\bibinfo{volume}{6}}, \bibinfo{pages}{291--306}
  (\bibinfo{year}{2024}).

\bibitem{ge2016multidimensional}
\bibinfo{author}{Ge, T.} \emph{et~al.}
\newblock \bibinfo{title}{Multidimensional heritability analysis of
  neuroanatomical shape}.
\newblock \emph{\bibinfo{journal}{Nature Communications}}
  \textbf{\bibinfo{volume}{7}}, \bibinfo{pages}{13291} (\bibinfo{year}{2016}).

\bibitem{primus2025beyond}
\bibinfo{author}{Primus, S.~A.} \emph{et~al.}
\newblock \bibinfo{title}{Beyond volume: Unraveling the genetics of human brain
  geometry}.
\newblock \emph{\bibinfo{journal}{Science Advances}}
  \textbf{\bibinfo{volume}{11}}, \bibinfo{pages}{eadr1644}
  (\bibinfo{year}{2025}).

\bibitem{naqvi2021shared}
\bibinfo{author}{Naqvi, S.} \emph{et~al.}
\newblock \bibinfo{title}{Shared heritability of human face and brain shape}.
\newblock \emph{\bibinfo{journal}{Nature Genetics}}
  \textbf{\bibinfo{volume}{53}}, \bibinfo{pages}{830--839}
  (\bibinfo{year}{2021}).

\bibitem{shadrin2021vertex}
\bibinfo{author}{Shadrin, A.~A.} \emph{et~al.}
\newblock \bibinfo{title}{Vertex-wise multivariate genome-wide association
  study identifies 780 unique genetic loci associated with cortical
  morphology}.
\newblock \emph{\bibinfo{journal}{NeuroImage}} \textbf{\bibinfo{volume}{244}},
  \bibinfo{pages}{118603} (\bibinfo{year}{2021}).

\bibitem{goovaerts2023joint}
\bibinfo{author}{Goovaerts, S.} \emph{et~al.}
\newblock \bibinfo{title}{Joint multi-ancestry and admixed {GWAS} reveals the
  complex genetics behind human cranial vault shape}.
\newblock \emph{\bibinfo{journal}{Nature Communications}}
  \textbf{\bibinfo{volume}{14}}, \bibinfo{pages}{7436} (\bibinfo{year}{2023}).

\bibitem{burns2024genetic}
\bibinfo{author}{Burns, R.} \emph{et~al.}
\newblock \bibinfo{title}{Genetic basis of right and left ventricular heart
  shape}.
\newblock \emph{\bibinfo{journal}{Nature Communications}}
  \textbf{\bibinfo{volume}{15}}, \bibinfo{pages}{9437} (\bibinfo{year}{2024}).

\bibitem{vandermeer2021genetic}
\bibinfo{author}{{van der Meer}, D.} \emph{et~al.}
\newblock \bibinfo{title}{The genetic architecture of human cortical folding}.
\newblock \emph{\bibinfo{journal}{Science Advances}}
  \textbf{\bibinfo{volume}{7}}, \bibinfo{pages}{eabj9446}
  (\bibinfo{year}{2021}).

\bibitem{fischl2012freesurfer}
\bibinfo{author}{Fischl, B.}
\newblock \bibinfo{title}{{FreeSurfer}}.
\newblock \emph{\bibinfo{journal}{NeuroImage}} \textbf{\bibinfo{volume}{62}},
  \bibinfo{pages}{774--781} (\bibinfo{year}{2012}).

\bibitem{miller2016multimodal}
\bibinfo{author}{Miller, K.~L.} \emph{et~al.}
\newblock \bibinfo{title}{Multimodal population brain imaging in the {UK
  Biobank} prospective epidemiological study}.
\newblock \emph{\bibinfo{journal}{Nature Neuroscience}}
  \textbf{\bibinfo{volume}{19}}, \bibinfo{pages}{1523--1536}
  (\bibinfo{year}{2016}).

\bibitem{xia2026sumh2}
\bibinfo{author}{Xia, T.} \emph{et~al.}
\newblock \bibinfo{title}{{Sum-h2}, enabling genetic discovery for deep
  learning-derived phenotypes through a fast evaluation framework and arena of
  performance}.
\newblock \emph{\bibinfo{journal}{bioRxiv}}  (\bibinfo{year}{2026}).
\newblock
  \urlprefix\url{https://www.biorxiv.org/content/10.64898/2026.09.24.753959v1}.
\newblock \bibinfo{note}{Preprint}.

\bibitem{radford2021learning}
\bibinfo{author}{Radford, A.} \emph{et~al.} \bibinfo{title}{Learning
  Transferable Visual Models From Natural Language Supervision}.
\newblock In \emph{\bibinfo{booktitle}{Proceedings of the 38th International
  Conference on Machine Learning}}, Vol. \bibinfo{volume}{139} of
  \emph{\bibinfo{series}{Proceedings of Machine Learning Research}},
  \bibinfo{pages}{8748--8763} (\bibinfo{year}{2021}).

\bibitem{xia2026learning}
\bibinfo{author}{Xia, T.} \emph{et~al.}
\newblock \bibinfo{title}{Learning heritable multimodal brain representation
  via contrastive learning}.
\newblock \emph{\bibinfo{journal}{bioRxiv}}  (\bibinfo{year}{2026}).
\newblock
  \urlprefix\url{https://www.biorxiv.org/content/10.64898/2026.02.19.706893v1}.
\newblock \bibinfo{note}{Preprint}.

\bibitem{vaswani2017attention}
\bibinfo{author}{Vaswani, A.} \emph{et~al.} \bibinfo{title}{Attention Is All
  You Need}.
\newblock In \emph{\bibinfo{booktitle}{Advances in Neural Information
  Processing Systems}}, Vol.~\bibinfo{volume}{30}, \bibinfo{pages}{5998--6008}
  (\bibinfo{year}{2017}).

\bibitem{chen2021crossvit}
\bibinfo{author}{Chen, C.-F.~R.}, \bibinfo{author}{Fan, Q.} \&
  \bibinfo{author}{Panda, R.} \bibinfo{title}{{CrossViT}: Cross-Attention
  Multi-Scale Vision Transformer for Image Classification}.
\newblock In \emph{\bibinfo{booktitle}{Proceedings of the IEEE/CVF
  International Conference on Computer Vision (ICCV)}},
  \bibinfo{pages}{357--366} (\bibinfo{year}{2021}).

\bibitem{deleeuw2015magma}
\bibinfo{author}{{de Leeuw}, C.~A.}, \bibinfo{author}{Mooij, J.~M.},
  \bibinfo{author}{Heskes, T.} \& \bibinfo{author}{Posthuma, D.}
\newblock \bibinfo{title}{{MAGMA}: Generalized Gene-Set Analysis of {GWAS}
  Data}.
\newblock \emph{\bibinfo{journal}{PLoS Computational Biology}}
  \textbf{\bibinfo{volume}{11}}, \bibinfo{pages}{e1004219}
  (\bibinfo{year}{2015}).

\bibitem{gtex2020atlas}
\bibinfo{author}{{GTEx Consortium}}.
\newblock \bibinfo{title}{The {GTEx} Consortium atlas of genetic regulatory
  effects across human tissues}.
\newblock \emph{\bibinfo{journal}{Science}} \textbf{\bibinfo{volume}{369}},
  \bibinfo{pages}{1318--1330} (\bibinfo{year}{2020}).

\bibitem{bulik2015atlas}
\bibinfo{author}{Bulik-Sullivan, B.}, \bibinfo{author}{Finucane, H.~K.},
  \bibinfo{author}{Anttila, V.}, \bibinfo{author}{Gusev, A.} \emph{et~al.}
\newblock \bibinfo{title}{An atlas of genetic correlations across human
  diseases and traits}.
\newblock \emph{\bibinfo{journal}{Nature Genetics}}
  \textbf{\bibinfo{volume}{47}}, \bibinfo{pages}{1236--1241}
  (\bibinfo{year}{2015}).

\bibitem{hawrylycz2012anatomically}
\bibinfo{author}{Hawrylycz, M.~J.}, \bibinfo{author}{Lein, E.~S.},
  \bibinfo{author}{Guillozet-Bongaarts, A.~L.}, \bibinfo{author}{Shen, E.~H.}
  \emph{et~al.}
\newblock \bibinfo{title}{An anatomically comprehensive atlas of the adult
  human brain transcriptome}.
\newblock \emph{\bibinfo{journal}{Nature}} \textbf{\bibinfo{volume}{489}},
  \bibinfo{pages}{391--399} (\bibinfo{year}{2012}).

\bibitem{barbeira2018exploring}
\bibinfo{author}{Barbeira, A.~N.}, \bibinfo{author}{Dickinson, S.~P.},
  \bibinfo{author}{Bonazzola, R.}, \bibinfo{author}{Zheng, J.} \emph{et~al.}
\newblock \bibinfo{title}{Exploring the phenotypic consequences of tissue
  specific gene expression variation inferred from {GWAS} summary statistics}.
\newblock \emph{\bibinfo{journal}{Nature Communications}}
  \textbf{\bibinfo{volume}{9}}, \bibinfo{pages}{1825} (\bibinfo{year}{2018}).

\bibitem{sudlow2015uk}
\bibinfo{author}{Sudlow, C.} \emph{et~al.}
\newblock \bibinfo{title}{UK Biobank: an open access resource for identifying
  the causes of a wide range of complex diseases of middle and old age}.
\newblock \emph{\bibinfo{journal}{PLoS medicine}}
  \textbf{\bibinfo{volume}{12}}, \bibinfo{pages}{e1001779}
  (\bibinfo{year}{2015}).

\bibitem{bycroft2018uk}
\bibinfo{author}{Bycroft, C.} \emph{et~al.}
\newblock \bibinfo{title}{The UK Biobank resource with deep phenotyping and
  genomic data}.
\newblock \emph{\bibinfo{journal}{Nature}} \textbf{\bibinfo{volume}{562}},
  \bibinfo{pages}{203--209} (\bibinfo{year}{2018}).

\bibitem{desikan2006automated}
\bibinfo{author}{Desikan, R.~S.}, \bibinfo{author}{S{\'e}gonne, F.},
  \bibinfo{author}{Fischl, B.}, \bibinfo{author}{Quinn, B.~T.} \emph{et~al.}
\newblock \bibinfo{title}{An automated labeling system for subdividing the
  human cerebral cortex on {MRI} scans into gyral based regions of interest}.
\newblock \emph{\bibinfo{journal}{NeuroImage}} \textbf{\bibinfo{volume}{31}},
  \bibinfo{pages}{968--980} (\bibinfo{year}{2006}).

\bibitem{watanabe2017functional}
\bibinfo{author}{Watanabe, K.}, \bibinfo{author}{Taskesen, E.},
  \bibinfo{author}{Van~Bochoven, A.} \& \bibinfo{author}{Posthuma, D.}
\newblock \bibinfo{title}{Functional mapping and annotation of genetic
  associations with FUMA}.
\newblock \emph{\bibinfo{journal}{Nature communications}}
  \textbf{\bibinfo{volume}{8}}, \bibinfo{pages}{1826} (\bibinfo{year}{2017}).

\end{thebibliography}
\endgroup

\clearpage
\section*{Supplementary Information}
\setcounter{table}{0}
\renewcommand{\thetable}{S\arabic{table}}

\begin{table}[!htbp]
\caption{\textbf{Notation.}}\label{tab:notation}
\centering
\small
\begin{tabularx}{\textwidth}{@{}lX@{}}
\toprule
Symbol & Meaning\\
\midrule
$x$, $\hat{x}$ & Input T1-weighted volume and its reconstruction\\
$z_{\mathrm{vol}}\in\mathbb{R}^{128}$ & Embedding of the T1 volume from the volumetric CNN encoder; the volume-only phenotypes\\
$z_{\mathrm{mesh}}\in\mathbb{R}^{128}$ & Embedding of the cortical surface from the frozen mesh encoder; the mesh-only phenotypes\\
$z'_{\mathrm{mesh}}$ & $z_{\mathrm{mesh}}$ after the learned adapter\\
$\delta\in\mathbb{R}^{128}$ & Surface-informed residual computed by the fusion module\\
$z_{\mathrm{fused}}=z_{\mathrm{vol}}+\delta$ & MEVA phenotypes\\
$D$ & Distillation head predicting $z_{\mathrm{mesh}}$ from $z_{\mathrm{fused}}$\\
$\mathcal{L}_{\mathrm{T1}}$, $\mathcal{L}_{\mathrm{mesh}}$ & T1 reconstruction loss and mesh-embedding distillation loss\\
$\lambda$ & Weight of $\mathcal{L}_{\mathrm{mesh}}$ ($\lambda=10$ in the main model)\\
\bottomrule
\end{tabularx}
\end{table}

\begin{table}[!htbp]
\caption{\textbf{Model capacity and training.}}\label{tab:capacity}
\small
\begin{tabular*}{\textwidth}{@{\extracolsep{\fill}}lll@{}}
\toprule
Module & Parameters & Training\\
\midrule
CNN autoencoder & 138.1M & 40 epochs (pretraining)\\
Mesh autoencoder & 2.91M & epoch 498 of 500\\
MEVA fusion modules & 0.18M & 4 + 6 epochs (two stages)\\
\botrule
\end{tabular*}
\footnotetext{The CNN autoencoder gives the volume-only phenotypes and initializes the MEVA volume encoder, which is fine-tuned during fusion; it has 69.6M encoder and 68.5M decoder parameters. The mesh autoencoder gives the mesh-only phenotypes and is frozen in MEVA. The MEVA fusion modules are the mesh adapter, the fusion block and the distillation head.}
\end{table}

\end{document}